\def\CleanBuild{1}
\PassOptionsToPackage{table}{xcolor}
\documentclass[]{style}
\usepackage[most]{tcolorbox}        % tcolorbox高级特性（加载完整库）
\usepackage{amsmath,amssymb}
\usepackage{lipsum}                 % 假文生成
\usepackage{xcolor}                 % 颜色
\usepackage{fancyhdr}               % 页眉（仅保留空样式）
\usepackage{booktabs}
\usepackage[table]{xcolor}
\usepackage{url}
\usepackage{graphicx}   % 用于插入图片
\usepackage{subcaption} % 用于生成子图和子标题
\usepackage{multirow}
\usepackage{bm}
\usepackage{adjustbox}  % 原内容稿附录逐任务大表需要（模板骨架补充）
\usepackage{ragged2e}   % 原内容稿附录表格 \RaggedRight 需要（模板骨架补充）
\DeclareUnicodeCharacter{2011}{-}   % 原内容稿附录含 U+2011 non-breaking hyphen，pdflatex 下映射为连字符
\definecolor{lightpurple}{RGB}{153, 102, 204}  % 马卡龙浅紫（好看且清晰）
\definecolor{lilac}{RGB}{182, 133, 210}        % 丁香紫（更柔和）
\newif\ifreviewversion
\ifdefined\ReviewBuild\reviewversiontrue\else\reviewversionfalse\fi
\ifdefined\CleanBuild\reviewversionfalse\fi
\newcommand{\rev}[1]{{\ifreviewversion\color{red}\fi#1}}
\usepackage{etoolbox}
\usepackage{hyperref}
\hypersetup{
    pdftitle={VISTA: Active Exploration for Self-Improvement in Embodied Agents},
    pdfauthor={Zhang Zhang, Shengjun Zhang, Simin Huang, Min Chen, Chong Xia, Xin Fei, Yueqi Duan},
    colorlinks=true,      % 启用彩色链接，而不是给链接加方框
    linkcolor=purple,        % 内部交叉引用的颜色（如图表、公式）
    citecolor=lilac,       % 参考文献引用的颜色（这里设为蓝色）
    urlcolor=magenta      % 网页 URL 的颜色
}
\newcommand{\VistaChangedBibKeys}{}
\forcsvlist{\listadd\VistaChangedBibKeys}{aspire2026,dgm2026,ding2026zetta,fu2026cap,gepa2026,hirobot2025,humanclaw,lei2026vlasastools,libero2023,lima2026agentic,lu2026aspire,metaharness2026,octo_2023,openx2023,palme2023,parallelworld2026,pi02024,pi052025,react2023,reflexion2023,robocasa2024,robotwin2025,rpent2026,rt22023,saycan2022,sayplan,shinn2023reflexion,wu2026lingbotvla,zetta2026}
\let\VistaOriginalBibitem\bibitem
\renewcommand{\bibitem}[2][]{%
  \color{black}%
  \if\relax\detokenize{#1}\relax\VistaOriginalBibitem{#2}\else\VistaOriginalBibitem[#1]{#2}\fi
  \color{black}%
  \ifreviewversion\ifinlist{#2}{\VistaChangedBibKeys}{\color{red}}{}\fi
}
\usepackage{tabularx}
\usepackage{lmodern}

\input{glyphtounicode}
\usepackage{tikz}
\usetikzlibrary{arrows.meta,positioning,calc,fit}
\usepackage{enumitem}
\usepackage{listings}
\newif\ifdraftreview
\draftreviewtrue

\newcommand{\TBD}[1]{\textcolor{thupurple}{\textsf{[#1]}}}
\newcommand{\SetSetting}[2]{\expandafter\def\csname ersiSetting#1\endcsname{#2}}
\newcommand{\SetResult}[2]{\expandafter\def\csname ersiResult#1\endcsname{#2}}
\newcommand{\Setting}[1]{\ifcsname ersiSetting#1\endcsname\csname ersiSetting#1\endcsname\else\TBD{#1}\fi}
\newcommand{\Result}[1]{\ifcsname ersiResult#1\endcsname\csname ersiResult#1\endcsname\else\TBD{#1}\fi}

\setlist[itemize]{leftmargin=1.4em,itemsep=0.15em,topsep=0.25em}
\setlist[enumerate]{leftmargin=1.7em,itemsep=0.15em,topsep=0.25em}
\microtypesetup{expansion=false}

\usepackage{amsmath,amssymb}
\usepackage{algorithm}
\usepackage{algpseudocode}
\usepackage{wrapfig}
\usepackage{needspace}

\definecolor{pwYellow}{RGB}{252,238,210}
\definecolor{pwPurple}{RGB}{239,220,247}
\definecolor{pwBlue}{RGB}{218,233,252}
\definecolor{pwGreen}{RGB}{218,241,234}

\definecolor{abstractpurple}{HTML}{9C27B0} % 摘要框紫色

\renewcommand{\maketitle}{\mymaketitle}

\begin{document}

% 论文内容
\title{
VISTA: Active Exploration for Self-Improvement in Embodied Agents
}
\author{Zhang Zhang$^{1,*}$, Shengjun Zhang$^{1,*, \dag}$, Simin Huang$^{1}$, Min Chen$^{1}$, Chong Xia$^{1}$, \\ Xin Fei$^{2}$, Yueqi Duan$^{1,\ddag}$}
\affiliation[]{$^{1}$Tsinghua University, $^{2}$National University of Singapore}

\begingroup
\renewcommand{\thefootnote}{}% 清除编号
\footnotetext{$*$ Equal contribution, $\dag$ Project Leader, $\ddag$ Corresponding author}% 输出无编号的脚注文本
\endgroup

% 摘要内容以 原内容稿为准；相对原内容稿的改动（补入 RoboTwin）用红色标出。
\abstract{
Agentic embodied intelligence plays a foundational role in open-ended, dynamic physical environments.
Rather than treating policy learning as an end-to-end mapping from instructions to actions, recent studies reveal that embodied systems depend not only on pre-trained foundation models, but also on harness.
Prior approaches largely treat rollouts as static logs for post-hoc reflection, producing
harness updates without autonomously testing and refining them through subsequent embodied interaction.
To address this challenge, we propose VISTA, a meta-agent framework that transforms post-hoc rollout reflection into active self-exploration for embodied harness improvement.
Specifically, VISTA maintains a harness graph that records the lineage of proposed harnesses, which enables the meta-agent to retrieve prior hypotheses, build on effective mechanisms, and avoid repeatedly exploring failed directions.
Each iteration begins from a retained harness version and its execution evidence. Rather than directly committing an update from these historical trajectories, the meta-agent actively explores in the environment.
It autonomously generates new rollouts, probes execution failures and recovery behaviors, and iteratively modifies the harness based on the resulting evidence. The refined harness is then evaluated afresh and added to the harness graph.
We evaluate VISTA on the LIBERO-Pro,  RoboCasa, and RoboTwin benchmarks.
Experimental results demonstrate that VISTA progressively improves task success and is more rollout-efficient than static reflection, achieving higher success rates with fewer simulator rollouts.
}

\checkdata[
\raisebox{-0.2em}{\includegraphics[width=0.025\linewidth]{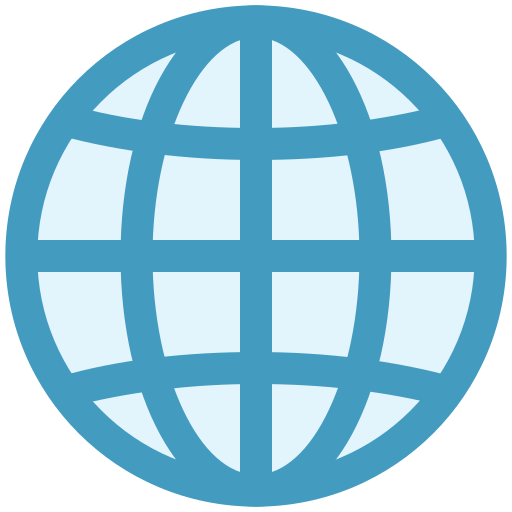}}~~Project Page]{\href{https://agenticei.github.io/VISTA-page/}{\texttt{agenticei.github.io/VISTA-page/}}
\\[-1.5ex]}

\checkdata[
\raisebox{-0.2em}{\includegraphics[width=0.025\linewidth]{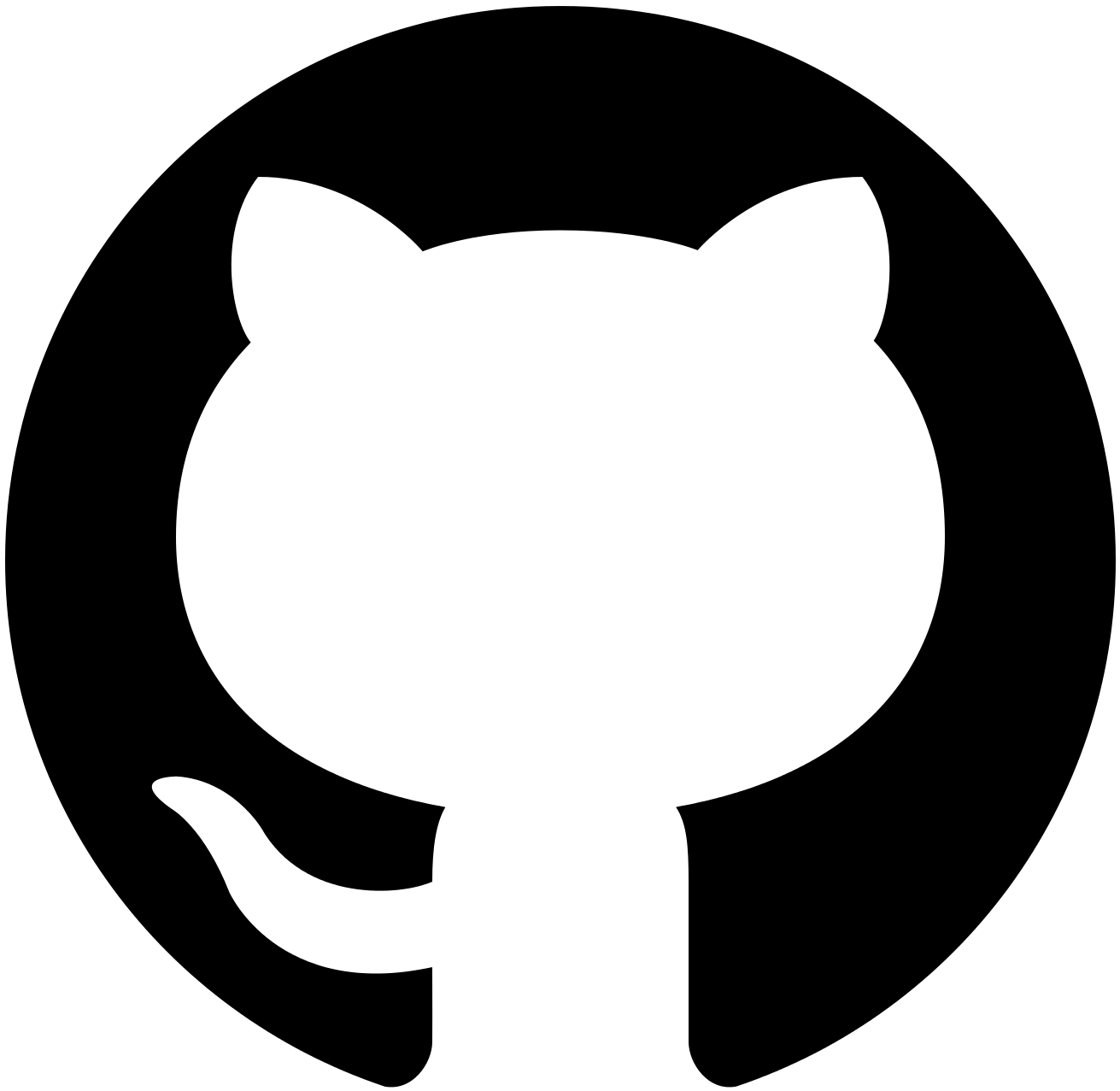}}~~GitHub Repo]{\href{https://github.com/AgenticEI/VISTA}{\texttt{github.com/AgenticEI/VISTA}}
\\[-1.5ex]}

% 调用标题+摘要（无报错）
\maketitle

\begin{figure}[h]
    \centering
    \includegraphics[width=0.95\linewidth]{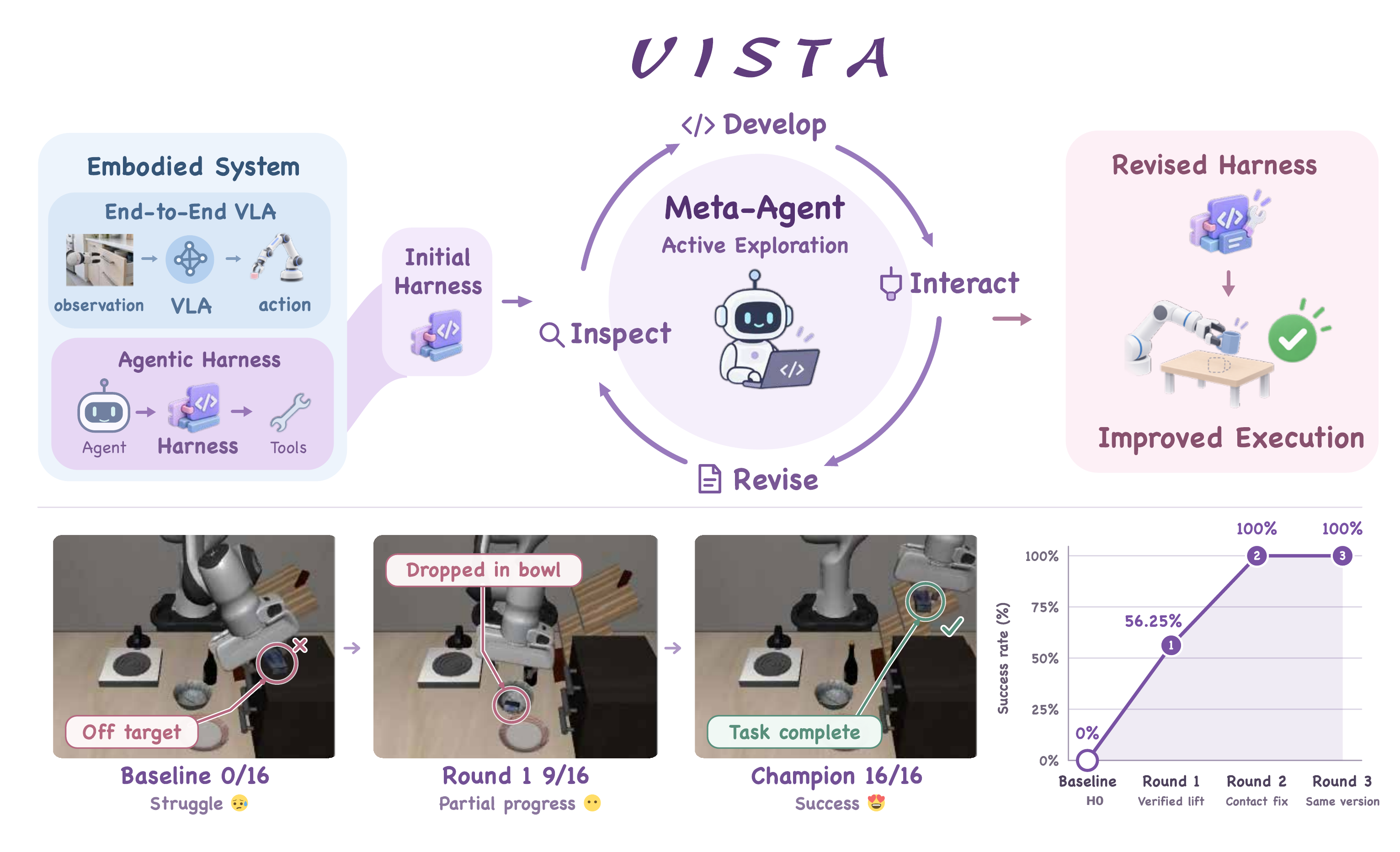}
    \vspace{-0.3cm}
    \caption{
    \textbf{VISTA actively explores the environment to improve embodied-agent harnesses.}
    Starting from an initial harness, a meta-agent inspects prior execution evidence, proposes revisions, and actively interacts with the environment to test hypotheses and refine candidate harnesses before formal evaluation.
    }
    \vspace{-0.5cm}
    \label{fig:placeholder}
\end{figure}

\section{Introduction}

Embodied artificial intelligence has emerged as a cornerstone for enabling autonomous systems to perceive, reason, and interact seamlessly within dynamic physical environments, offering transformative potential across household assistance, healthcare logistics, and industrial automation~\citep{brohan2022rt1,openx2023}. Catalyzed by multimodal foundation models and large-scale embodiment datasets, learning-based paradigms have demonstrated increasingly competitive capabilities in executing atomic manipulation tasks~\citep{rt22023,octo_2023,openvla2024,pi02024,pi052025,wu2023gr1,lyu2026lda,ye2026dreamzero,li2026lingbotva}.
But scaling end-to-end policies to long-horizon, multi-stage tasks remains notoriously difficult. End-to-end models suffer severely from compounding execution errors over extended trajectories, out-of-distribution generalization under perturbations, and prohibitive sample complexity when attempting to fit multi-modal, long-tail action distributions across heterogeneous task phases~\citep{ahn2022can,innermonologue2022}.

To overcome these fundamental limitations, the convergence of frontier large language model (LLM) agents and robotics has established embodied agentic systems as a pivotal research frontier~\citep{codeaspolicies2023,singh2022progprompt,voxposer2023,voyager2023,rpent2026}. By coupling high-level symbolic reasoning and stateful tool orchestration with low-level control policies, agentic embodied architectures offer decisive advantages in managing long-horizon complexity: they facilitate modular hierarchical task decomposition, support active perception to resolve visual ambiguities, and enable closed-loop error diagnosis and autonomous recovery.
The performance, reliability, and long-horizon coherence of such systems depend not only on their pre-trained foundation model weights, but also fundamentally on their harness~\citep{yang2024swe,reflexion2023}.
\rev{Existing systems} construct \rev{and improve these} harnesses \rev{through code generation}, \rev{reusable skills, and memory-guided execution~}\rev{\citep{codeaspolicies2023,singh2022progprompt,voyager2023,rpent2026}}\rev{. VISTA focuses} on \rev{a complementary question: how can a coding agent improve an executable robot harness by acquiring additional evidence through direct interaction with the environment? Existing records do not always contain the observations} or \rev{interventions needed to test a proposed explanation of a failure}.

To address this issue, we propose \textbf{VISTA}, a meta-agent framework that unifies active environmental exploration with closed-loop harness adaptation to achieve embodied self-improvement. \rev{The meta-agent controls the development process: it selects what to inspect, which public sensing and control capabilities to invoke, and which autonomous tests to request. Evidence from these interactions guides executable} harness \rev{revisions.
After development}, the \rev{submitted harness is frozen} and \rev{evaluated through fresh autonomous execution. Its code and} evidence are \rev{retained in a harness DAG}, \rev{allowing the} meta-agent to \rev{retrieve} prior hypotheses, reuse effective mechanisms, and \rev{explore} from earlier candidates.

We evaluate VISTA on LIBERO-Pro~\rev{\cite{liberopro2025}}\rev{, RoboCasa365~}\rev{\cite{robocasa3652026}}\rev{,} and \rev{RoboTwin~}\rev{\cite{robotwin2025}}.
VISTA improves task success \rev{by accumulating executable mechanisms across development rounds}. These results support \rev{harness development} as \rev{a practical way to convert} embodied experience into \rev{persistent changes} in \rev{agent behavior}.
Our contributions are summarized as follows:
\begin{itemize}
    \item  We propose VISTA, a framework that departs from conventional handcrafted heuristics and static memory augmentation by framing the robot's operational harness as an evolvable, executable control plane, realizing autonomous embodied self-improvement.
    \item We introduce an active exploration mechanism where the meta-agent empowers the embodied system to actively gather diagnostic physical evidence, \rev{linking} exploratory \rev{evidence to executable harness revisions}. Furthermore, we develop a Harness Graph to track the evolutionary lineage of candidate harnesses, facilitating hypothesis retrieval, component reuse, and principled branch exploration.
    \item We conduct comprehensive evaluations on the challenging LIBERO-Pro\rev{, RoboCasa,} and \rev{RoboTwin} benchmarks. Experimental results demonstrate that VISTA progressively improves manipulation success rates through accumulated experience, \rev{improving over the corresponding initial harnesses}.
\end{itemize}

\section{Related Work}

\label{sec:related-work}

\subsection{Foundation Models for Robotic Manipulation}

\paragraph{Low-level action policies.}
Foundation models for robotic manipulation can be broadly categorized according to how their semantic knowledge is connected to physical control. One line of research extends pretrained vision--language models (VLMs) into low-level policies through vision--language--action (VLA) models, which generate continuous action chunks using diffusion or flow matching \citep{pi02024,diffusionpolicy}, task-space keyframes or spatial-grid actions \citep{spatialvla}, and latent action codes learned from robot trajectories or videos \citep{univla,moto}. Recent video--action and world--action models further couple action prediction with future visual dynamics \citep{lightwam,selfwam}. Robot foundation models extend these capabilities across heterogeneous tasks and embodiments \citep{scaling,openx2023}, while subsequent studies improve action initialization, tokenization, and adaptation to new tasks and embodiments \citep{apt}. Despite their strong control performance, these approaches translate general pretrained knowledge into embodiment-specific motor signals and often require additional robot trajectories when the task family or platform changes \citep{zhong2025survey}.

\paragraph{Semantic decision making and physical grounding.}
A complementary line of research treats foundation models as semantic decision makers rather than direct motor controllers. These models predict subgoals, keypoints, affordances, value maps, or spatial constraints, which are subsequently realized by downstream controllers \citep{palme2023,voxposer2023}. Although this preserves high-level semantic reasoning, it separates model intent from physical execution and depends on system-specific grounding pipelines \citep{matters,manipulate}, as well as long-horizon planning, closed-loop feedback, and failure recovery \citep{addressing}. Together, these two directions reveal a trade-off between direct physical control and semantic generalization: low-level policies offer precise execution but are closely tied to specific embodiments, whereas intermediate representations preserve semantic structure but limit physical adaptability. This trade-off motivates interfaces that combine semantic interpretability with fine-grained executable actions.

\subsection{Embodied Agent Systems}

\paragraph{Agentic control architectures.}
Building on these foundation-model paradigms, agentic embodied systems place a pretrained VLM within an iterative harness that translates model decisions into robot actions and feeds execution outcomes back to the model, thereby closing the perception--decision--execution loop \citep{lima2026agentic,guava}. Existing systems mainly differ in the primitives through which decisions are executed. Some models generate programs over perception and control APIs \citep{roboscript,fu2026cap}, while others select skills from predefined libraries \citep{roboclaw,sayplan}, steer learned policies such as VLAs with language subgoals \citep{volo,cortex}, or delegate decisions to symbolic planners and numerical optimizers \citep{autotamp,chen2025code}. To support long-horizon interaction, these systems typically maintain task context and persistent memory \citep{remembr}, perform feedback-based monitoring and reflection \citep{innermonologue2022}, and detect failures for replanning and recovery \citep{aha}.

\paragraph{Interfaces for embodied execution.}
Within agentic robot systems, interface design strongly affects control performance. While high-level, human-designed abstractions can improve reliability, composing low-level perception and control APIs remains difficult \citep{fu2026cap}. Many systems therefore expose skills, controllers, or VLAs as callable primitives \citep{shi2025hi,rpent2026}, leaving the agent to decide what to do while an opaque executor determines how to perform it \citep{addressing}. Although this simplifies planning, it reduces the model's role in fine-grained physical control. Similar compact action interfaces are used by computer-use and game agents \citep{showui,gameworld}, simulated embodied agents \citep{behavior,humanclaw}, and navigation systems \citep{zhou2026embodied}. \rev{These interfaces make model decisions executable; VISTA studies how the surrounding harness can itself improve through experience.}

\paragraph{\rev{Harness and skill improvement.}}
\rev{Meta-Harness optimizes harness code using prior implementations} and \rev{execution evidence~}\rev{\citep{metaharness2026}}\rev{. In robotics, ASPIRE develops} reusable \rev{skills through iterative exploration~}\rev{\citep{aspire2026}}, \rev{while Zetta evolves runtime critics and recovery skills~}\rev{\citep{zetta2026}}\rev{. VISTA connects Meta-controlled exploration to executable harness revisions and preserves their code, evidence}, and \rev{lineage for subsequent development}.

\section{Method}

\label{sec:method}

\begin{figure}[t]
    \centering
    \includegraphics[width=0.95\linewidth]{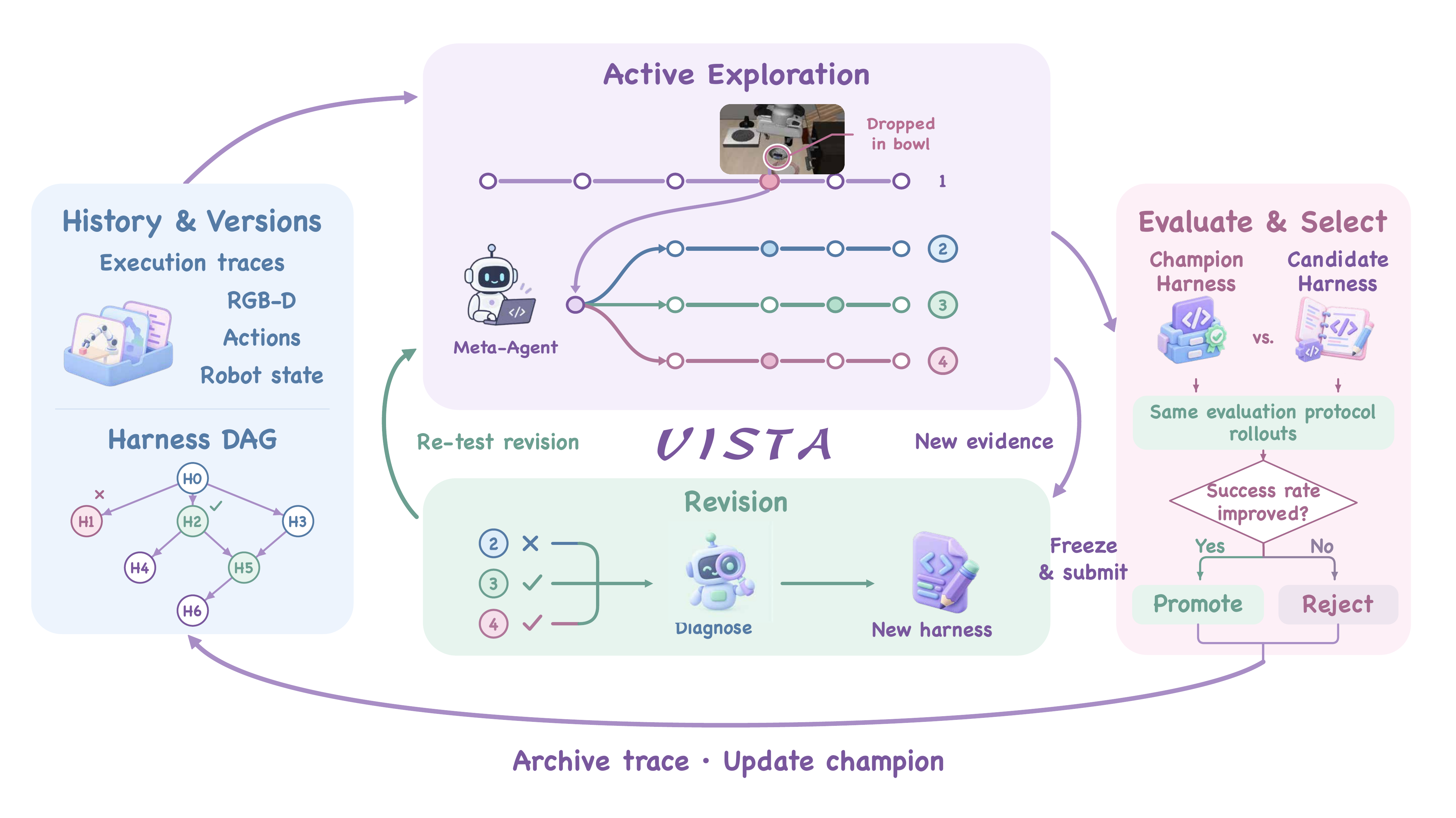}
    \caption{\textbf{Overview of VISTA.} The meta-agent develops an embodied
    harness from versioned implementations and multimodal execution
    evidence. Within a development iteration, it interleaves code revision,
    active exploration, and autonomous candidate \rev{testing}. After
    submission, the frozen candidate undergoes \rev{a separate} autonomous
    evaluation under the declared development conditions. The selection
    rule determines whether it replaces the incumbent; both promoted and
    rejected versions remain in the history. The inner loop acquires
    evidence and refines a candidate, while the outer loop selects and
    retains harnesses across development iterations.}
    \label{fig:pipeline}
\end{figure}

VISTA improves an embodied agent by evolving its harness, including the instructions, executable skills, and coordination logic through which the agent perceives, reasons, and acts. The online agent has access to a language-model planner, a VLA model, and public robot capabilities. VISTA keeps the underlying model parameters fixed and optimizes how these resources are composed into autonomous behavior. Figure~\ref{fig:pipeline} illustrates the resulting development loop. A meta-agent, implemented as a coding agent, retrieves a harness version and its execution evidence, identifies a behavior to improve, and develops a candidate. During development, it can actively investigate the environment, probe execution failures, and revise the harness in response to newly acquired evidence. It can also request autonomous tests of intermediate versions. The submitted candidate is then frozen and evaluated in fresh autonomous episodes. A host-controlled selection rule updates the incumbent, while the candidate and its evidence remain available for subsequent iterations. Improvement is thus carried forward through both executable harness versions and the experience accumulated while developing them.

\subsection{Harness DAG}
\label{sec:method:formulation}

Let $L_{\theta}$ denote the online language model, $\pi_{\phi}$ the VLA,
and $\mathcal{C}$ the public capability interface. The interface exposes
observation, perception, policy inference, and bounded robot control.
A harness is a package
$H=(I,\mathcal{S},P,M,D)$, where $I$ specifies model-facing instructions,
$\mathcal{S}$ contains skill descriptions and executable implementations,
$P$ provides hooks for request construction, tool selection, and feedback
processing, $M$ stores persistent procedural knowledge, and $D$ declares
tool bindings and capability dependencies. Skills may compose multiple
perception or control operations, while policy hooks govern their use
within the host's execution loop. Harness revisions can therefore change
both the operations available to the agent and how it coordinates them.
The model parameters, capability interface, and evaluation rule are fixed.
For a task instance $x$, comprising a goal and scene initialization, and
execution randomness $\xi$, let $S(H;x,\xi)\in\{0,1\}$ denote official
autonomous task success. Our objective is
\begin{equation}
    \max_{H\in\mathcal{H}_{\mathcal{C}}} J(H),
    \qquad
    J(H)=\mathbb{E}_{x\sim\mathcal{T},\,\xi}
    \bigl[S(H;x,\xi)\bigr],
    \label{eq:vista_objective}
\end{equation}
where $\mathcal{H}_{\mathcal{C}}$ is the space of admissible harnesses and
$\mathcal{T}$ is the task-instance distribution within the declared search
scope. A frozen harness still induces closed-loop behavior: its code and
persistent knowledge remain fixed during evaluation, while episode-local
state and actions depend on incoming observations.

Harness development produces a directed acyclic graph (DAG) $G=(V,E)$ of
immutable versions. We instantiate each harness as a Git-versioned
package, with commits identifying snapshots and parent links defining
lineage. Branches preserve alternative development directions, and explicit
merges can combine histories. Execution records and development notes are
linked to their corresponding versions. Each iteration begins from the
incumbent by default, but the meta-agent may select another recorded base,
including a previously rejected candidate, and inspect earlier
implementations for reusable mechanisms. Retaining unsuccessful proposals
makes their diagnostic findings available to later development. 

\subsection{Evidence-Guided Harness Revision}
\label{sec:method:evidence}

Each development record is tied to a frozen harness version and exact
execution condition, including task initialization, repetition, randomness,
and execution mode. A trajectory aligns planner decisions and tool calls with
robot actions, images, and state, allowing the meta-agent to localize a failure
from an aggregate outcome to a specific decision or physical execution interval.

Given a base harness and these records, the meta-agent forms a concrete
hypothesis and revises the implicated components, such as instructions,
observation processing, skill termination, or coordination logic. The resulting
version and its supporting evidence are retained for later development.
When recorded trajectories are insufficient to resolve the hypothesis, VISTA
acquires additional embodied evidence through active exploration.

\subsection{Active Diagnostic Exploration}
\label{sec:method:active}

Unlike static reflection, VISTA allows the meta-agent to collect new embodied
evidence before submitting a candidate. It can launch a private simulator
session from a frozen harness snapshot, inspect the current execution state,
and invoke public sensing or control capabilities to test a specific
hypothesis. These interactions provide diagnostic evidence for subsequent
harness revisions rather than modifying the formal evaluation process.

After revision, the meta-agent can request fresh autonomous test rollouts to
determine whether the candidate reproduces the desired behavior without
intervention. Development therefore alternates among hypothesis formation,
targeted interaction, harness revision, and autonomous testing until a
candidate is submitted or the development budget is exhausted. Diagnostic
traces and intermediate test results remain development evidence; promotion is
determined only by the separate formal evaluation in \ref{sec:method:selection}.

\subsection{Autonomous Validation and Selection}
\label{sec:method:selection}

After submission, the host evaluates the frozen candidate through fresh
episodes without meta-agent intervention. For a declared development
matrix containing $M$ task initializations and $R$ repetitions per
initialization, its empirical success is
\begin{equation}
    \widehat{J}_{\mathrm{dev}}(H)
    =\frac{1}{MR}\sum_{i=1}^{M}\sum_{r=1}^{R}
      S(H;x_i,\xi_{i,r}).
    \label{eq:vista_dev_score}
\end{equation}
Candidate and incumbent evaluations use matching task initializations,
repetition settings, model configurations, and execution limits. Selection
requires a complete, comparable matrix of valid autonomous outcomes.
Matching incumbent evaluations may be reused. Under the success-based
selection rule, the incumbent $H_k^{\star}$ is updated as
\begin{equation}
    H_{k+1}^{\star}=
    \begin{cases}
        \widetilde H_k,
        &\widehat{J}_{\mathrm{dev}}(\widetilde H_k)
         >\widehat{J}_{\mathrm{dev}}(H_k^{\star}),\\
        H_k^{\star},&\text{otherwise},
    \end{cases}
    \label{eq:vista_selection}
\end{equation}
where $\widetilde H_k$ is the submitted candidate. Ties retain the
incumbent. All frozen versions and their available records are retained,
regardless of promotion. 

Algorithm~\ref{alg:vista} summarizes the procedure. Search is bounded by
the configured development and iteration budgets. When early stopping is
enabled, the same selected package must achieve the required consecutive
perfect formal evaluations; a confirmation iteration re-evaluates that
package without another development attempt. 

\begin{algorithm}[t]
    \caption{VISTA: active exploration for harness evolution}
    \label{alg:vista}
    \begin{algorithmic}[1]
        \Require Initial harness $H_{\mathrm{init}}$,
        development matrix $\mathcal{D}_{\mathrm{dev}}$, budgets

        \State Evaluate $H_{\mathrm{init}}$ and initialize
        $H^\star \gets H_{\mathrm{init}}$ and development graph $G$

        \While{budget remains and stopping rule is not met}
            \If{$H^\star$ requires confirmation}
                \State $\widetilde{H} \gets H^\star$
            \Else
                \State Initialize development from a retained base in $G$
(default: $H^\star$)
                \While{development budget remains and no candidate is submitted}
                    \State Meta-agent autonomously selects and executes the next
                    operation from evidence
                    \Statex \hspace{\algorithmicindent}
                    (history inspection, harness revision,
                    live diagnosis, or autonomous testing)
                    \State Add resulting versions and evidence to $G$
                \EndWhile
                \State Validate and freeze the submitted candidate
                $\widetilde{H}$
            \EndIf

            \State Evaluate $\widetilde{H}$ autonomously on
            $\mathcal{D}_{\mathrm{dev}}$ from fresh initializations
            \State Add $\widetilde{H}$ and its evaluation records to $G$
            \If{the evaluation is complete}
                \State Update $H^\star$ using Eq.~\eqref{eq:vista_selection}
                and update its confirmation state
            \EndIf
        \EndWhile

        \State \Return $H^\star$
    \end{algorithmic}
\end{algorithm}

\section{Experiments}

\label{sec:experiments}

\subsection{Experimental Setup}

\textbf{Benchmarks.} We conduct experiments on LIBERO-Pro~\citep{liberopro2025}\rev{, RoboCasa365~}\rev{\citep{robocasa3652026}}\rev{,} and \rev{RoboTwin~}\rev{\citep{robotwin2025}}. For LIBERO-Pro, we evaluate \rev{the} \textit{Spatial}, \textit{Object}, \textit{Goal}, and \textit{\rev{LIBERO-10}} \rev{suites} under \rev{both instruction-redirection (T) and position-swap (S) perturbations}. Each suite contains 10 tasks\rev{, evaluated using eight scene seeds with two repetitions per seed}. For RoboCasa365, we consider the \textit{Composite Seen} and \textit{Composite Unseen} \rev{categories, each containing} 16 tasks \rev{evaluated using eight scene seeds with two repetitions. For RoboTwin}, \rev{we evaluate 50} tasks \rev{in four randomized scenes with two repetitions per scene}.

\textbf{Models \rev{and capabilities}.} We use $\pi_{0.5}$~\rev{\citep{pi052025}}\rev{, RLDX-1~}\rev{\citep{rldx2026}}\rev{, and LingBot-VLA~}\rev{\citep{wu2026lingbotvla}} as the \rev{learned policies} for LIBERO-Pro\rev{, RoboCasa365,} and \rev{RoboTwin, respectively}. \rev{The} embodied planner \rev{uses} GPT-5.6 Sol, and the \rev{Meta-agent} is implemented \rev{with} Codex and GPT-6\rev{. Across the three benchmarks}, \rev{the public capability interface combines learned policies with perception} and \rev{native robot control}. \rev{For example, RoboTwin provides motion-planned end-effector movement, wrist rotation, and gripper control. Episodes allow up to 100 planner turns; LIBERO-Pro and RoboTwin use a 10,000-step environment horizon, while RoboCasa365 follows} the per-task \rev{horizons of} 300--2900 steps \rev{used by RLDX-1. Model identifiers} and \rev{launch configurations are provided in the code repository}.

\textbf{Baselines \rev{and initial harnesses}.} We compare \rev{VISTA} with end-to-end VLA models, including OpenVLA~\citep{openvla2024}, $\pi_0$~\citep{pi02024}, $\pi_{0.5}$~\citep{pi052025}, RLDX-1~\citep{rldx2026}, and \rev{LingBot-VLA~}\rev{\citep{wu2026lingbotvla}}, \rev{as well as} HarnessVLA~\rev{\citep{rpent2026}}. For LIBERO-Pro, our \rev{initial} harness $H_0$ \rev{exposes} the \rev{frozen} VLA and a \rev{minimal observation interface}. For RoboCasa365 \rev{and RoboTwin}, we \rev{construct $H_0$ by adapting} the HarnessVLA \rev{reference implementation to each benchmark}, \rev{retaining its} prompts, \rev{policy hooks,} tool \rev{interfaces}, and \rev{memory. VISTA develops these initial harnesses and selects a version for each task. Every evaluated candidate is frozen and executed autonomously}.

\begin{table}[t]
  \centering
  \scriptsize
  \setlength{\tabcolsep}{3pt}
  \caption{\textbf{Success rate (\%) on \rev{the eight} LIBERO-Pro \rev{settings}}. $H_0$ is the initial harness and $H_{\mathrm{VISTA}}$ is the selected harness. \rev{$\Delta$ is measured in percentage points.}}
  \label{tab:libero-main}
  \begin{adjustbox}{max width=\linewidth}
  \begin{tabular}{lrrrrrrrrr}
    \toprule
    Method & \rev{Spatial-T} & \rev{Spatial-S} & \rev{Object-T} & \rev{Object-S} & \rev{Goal-T} & \rev{Goal-S} & \rev{LIBERO-10-T} & \rev{LIBERO-10-S} & \rev{Mean} \\
    \midrule
    OpenVLA~\citep{openvla2024} & 0.0 & \rev{0.0} & 0.0 & \rev{0.0} & 0.0 & \rev{0.0} & 0.0 & \rev{0.0} & \rev{0.0} \\
    $\pi_0$~\citep{pi02024} & 0.0 & \rev{0.0} & 0.0 & \rev{2.0} & 0.0 & \rev{0.0} & 0.0 & \rev{0.0} & \rev{0.3} \\
    $\pi_{0.5}$~\citep{pi052025} & \rev{1.0} & \rev{20.0} & \rev{1.0} & \rev{17.0} & \rev{2.0} & \rev{38.0} & \rev{1.0} & \rev{8.0} & \rev{11.0} \\
    HarnessVLA~\rev{\citep{rpent2026,parallelworld2026}} & 79.0 & \rev{56.0} & 79.0 & \rev{85.0} & 62.0 & \rev{57.0} & \rev{42.0} & \rev{29.0} & \rev{61.1} \\
    \midrule
    Initial harness $H_0$ & 19.0 & \rev{30.0} & 49.0 & \rev{14.4} & 16.0 & \rev{4.4} & 7.0 & \rev{0.0} & \rev{17.5} \\
    \rev{Evolved} harness $H_{\mathrm{VISTA}}$ & 98.0 & \rev{100.0} & 100.0 & \rev{99.4} & 99.0 & \rev{97.5} & 89.0 & \rev{76.9} & \rev{95.0} \\
    \midrule
    $\Delta$ over $H_0$ & +79.0 & \rev{+70.0} & +51.0 & \rev{+85.0} & +83.0 & \rev{+93.1} & +82.0 & \rev{+76.9} & \rev{+77.5} \\
    \bottomrule
  \end{tabular}
  \end{adjustbox}
\end{table}

\begin{table}[t]
  \centering
  \small
  \caption{\textbf{Success rate (\%) on Composite Seen and Composite Unseen \rev{tasks} in RoboCasa365.} $H_0$ is \rev{initialized} from HarnessVLA. \rev{$\Delta$ is measured in percentage points.}}
  \label{tab:robocasa-main}
  \begin{tabular}{lcc}
    \toprule
    Method & Composite Seen & Composite Unseen \\
    \midrule
    $\pi_{0.5}$ & 6.8 & 0.9 \\
    RLDX-1 & 17.4 & 6.8 \\
    \midrule
    Initial harness $H_0$ & \rev{16.4} & 4.7 \\
    \rev{Evolved} harness $H_{\mathrm{VISTA}}$ & \rev{24.2} & \rev{13.3} \\
    \midrule
    $\Delta$ over $H_0$ & +7.8 & +\rev{8.6} \\
    \bottomrule
  \end{tabular}
\end{table}

\begin{table}[t]
  \centering
  \ifreviewversion\color{red}\fi
  \small
  \setlength{\tabcolsep}{8pt}
  \caption{\textbf{Success rate (\%) on RoboTwin.}}
  \label{tab:robotwin-main}
  \begin{tabular}{lccccc}
    \toprule
    & \multicolumn{3}{c}{Baselines~\citep{rpent2026}} & \multicolumn{2}{c}{Harness evolution} \\
    \cmidrule(lr){2-4}\cmidrule(lr){5-6}
    Method & $\pi_{0.5}$ & LingBot-VLA & HarnessVLA & $H_0$ & $H_{\mathrm{VISTA}}$ \\
    \midrule
    Success rate (\%) & 47.9 & 50.4 & 58.0 & 48.5 & \textbf{74.0} \\
    \bottomrule
  \end{tabular}
\end{table}

\subsection{Main Results}

\textbf{Overall Performance.} \rev{Table~}\rev{\ref{tab:libero-main}} \rev{reports results} across \rev{all eight LIBERO-Pro settings}. \rev{VISTA improves the overall} success rate \rev{from $17.5\%$ to $95.0\%$, a gain} of \rev{77.5 percentage points over $H_0$. The evolved harnesses achieve }$97.0\%$ \rev{on the T settings} and \rev{$93.4\%$ on} the \rev{S settings, showing improvements under both instruction and position perturbations}. On the multi-stage household tasks of RoboCasa365, $H_{\mathrm{VISTA}}$ improves success from $16.4\%$ to \rev{$24.2\%$} on Composite Seen and from $4.7\%$ to \rev{$13.3\%$} on Composite Unseen \rev{(Table~}\rev{\ref{tab:robocasa-main}}\rev{). On RoboTwin}, \rev{success increases from $48.5\%$ to $74.0\%$ (Table~}\rev{\ref{tab:robotwin-main}}\rev{)}.

\begin{figure}[htbp]
  \centering
  \includegraphics[width=0.6\linewidth]{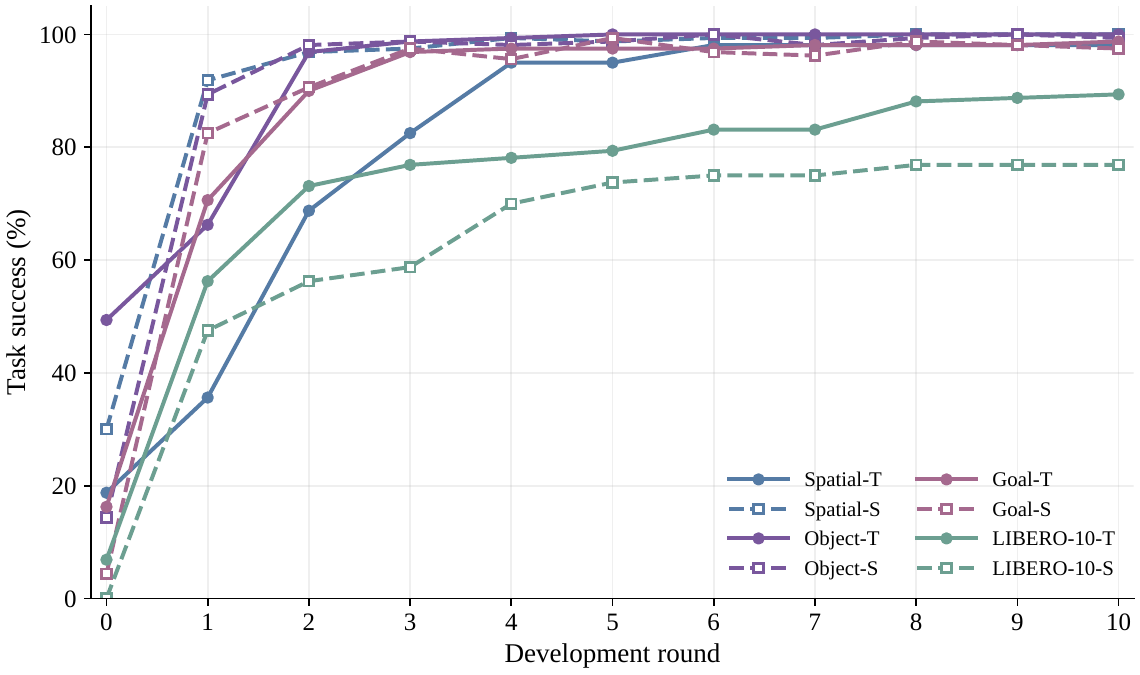}
  \caption{\textbf{Evolution progress of success rate (\%) on LIBERO-Pro.} \rev{Colors distinguish the four task} suites\rev{; circles denote T and squares denote S}.}
  \label{fig:libero-evolution}
\end{figure}

\textbf{Evolution over Development.} Figure~\ref{fig:libero-evolution} visualizes how autonomous success changes over development rounds. Each curve \rev{averages the selected-harness success rates} over \rev{the ten} tasks in one LIBERO-Pro \rev{setting}. \rev{The Spatial, Object, and Goal settings improve rapidly during} the \rev{early} rounds, while \rev{LIBERO-10 benefits from further development} over \rev{a longer sequence of revisions}.

\begin{table}[t]
  \centering
  \small
  \caption{\textbf{Comparison of passive and active exploration on four LIBERO-Goal tasks.} $\Delta$ is the mean improvement in success rate (percentage points) of the selected harness over the \rev{initial} $\pi_{0.5}$\rev{-based harness $H_0$}; Token is the mean cumulative Meta tokens per task after each round.}
  \label{tab:passive_active}
  \begin{tabular}{lcccc}
    \toprule
    \rev{Methods} & \multicolumn{2}{c}{\rev{$R_1$}} & \multicolumn{2}{c}{\rev{$R_2$}} \\
    \cmidrule(lr){2-3}\cmidrule(lr){4-5}
    & \rev{$\Delta$} & \rev{Token} & \rev{$\Delta$} & \rev{Token} \\
    \midrule
    Passive & +39.1 & 94.2M & +56.3 & 137.7M \\
    Active & \textbf{+50.0} & \textbf{42.2M} & \textbf{+85.9} & \textbf{91.6M} \\
    \bottomrule
  \end{tabular}
\end{table}

\begin{figure*}[t]
\centering
\includegraphics[width=\linewidth]{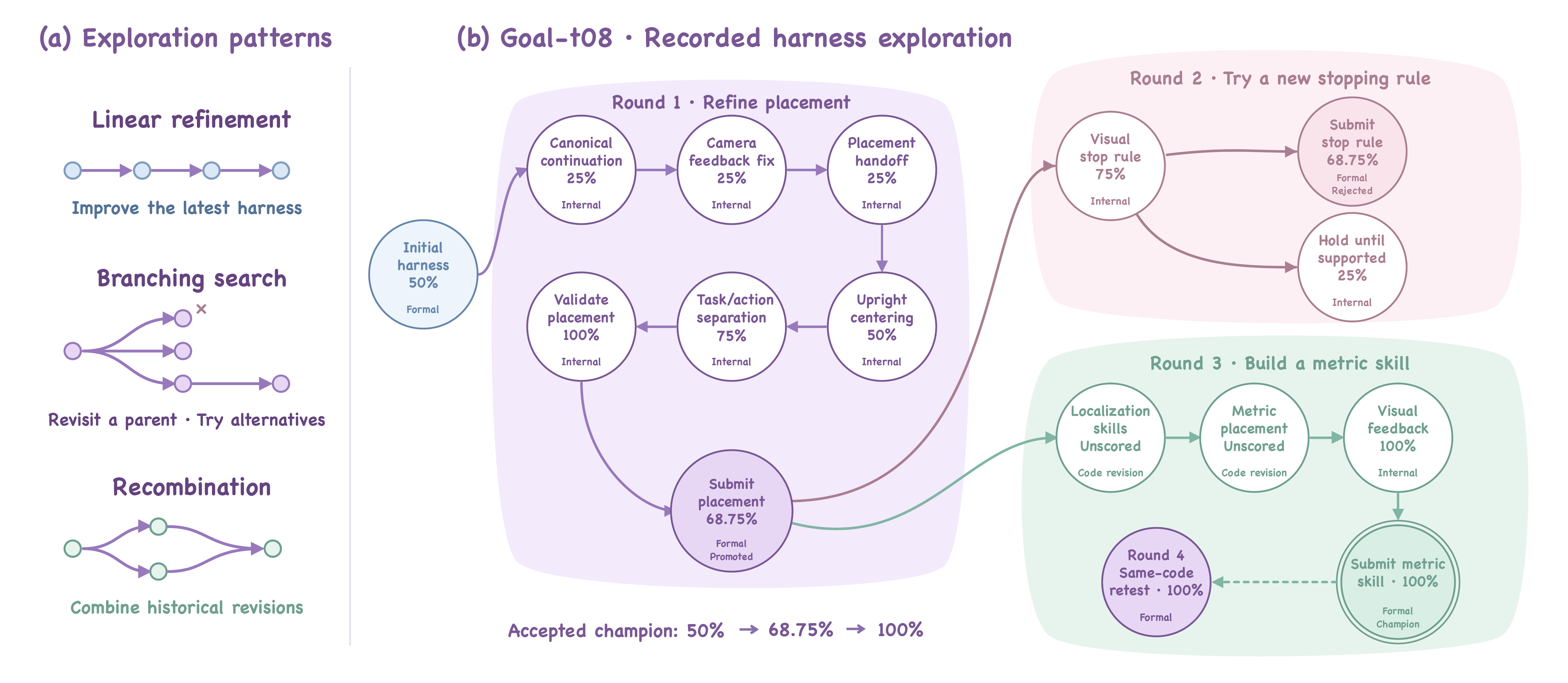}
\caption{\textbf{Harness development trajectories.} (a) Representative development patterns, including linear refinement, branching exploration, and branch recombination. (b) A trajectory on LIBERO Goal-t08. \emph{Internal} nodes denote small-scale exploration and \rev{testing} during development, \rev{while} \emph{formal} nodes \rev{evaluate frozen candidates} on the complete \rev{task} matrix and determine \rev{promotion}. \rev{Success} improves from \rev{50.0}\% to \rev{100.0}\%.}
\label{fig:development-tree}
\end{figure*}

\Needspace{7\baselineskip}
\subsection{Ablation}

\paragraph{Active vs.\ passive exploration.}
To \rev{assess} the contribution of active exploration, we compare VISTA with a passive, rollout-only variant \rev{on four LIBERO-Goal tasks}. The passive variant retains history inspection, harness editing, and autonomous candidate evaluation, but cannot open a live scene, probe a decision boundary, or execute diagnostic capabilities during development; it must diagnose failures from recorded rollouts alone. Meta tokens include cached and uncached \rev{input, output,} and \rev{sub-agent usage}.

Table~\ref{tab:passive_active} shows \rev{larger gains with} active exploration. After the first round, active exploration improves success by 50.0 \rev{percentage} points using 42.2M Meta tokens per task, whereas the passive variant improves by 39.1 points while consuming 94.2M. \rev{After two rounds,} the \rev{gains reach} 85.9 \rev{and} 56.3 \rev{percentage} points\rev{, respectively, with mean cumulative usage of 91.6M and 137.7M} tokens. \rev{Active exploration provides additional observations and interventions that help} the \rev{Meta-agent test its hypotheses and develop effective harness changes}.

\subsection{Analysis: Inside Harness Evolution}

\textbf{Development Trajectories.}
During optimization, the Meta Agent maintains a Git-like lineage to explore and revise harnesses. We analyze the recorded lineages to characterize how harnesses evolve across successive revisions. Depending on the task, we observe different development patterns. In some cases, optimization proceeds linearly by refining the current harness. In others, the optimizer branches from an earlier revision when the current hypothesis appears ineffective and explores an alternative direction. We also observe cases where mechanisms developed in different branches are recombined into a new harness.
Figure~\ref{fig:development-tree}(b) shows a representative trajectory on
LIBERO Goal-t08. Starting from an initial harness with 50\% formal success,
the first development round refines the placement behavior and promotes a
candidate with 68.75\% success. The next round explores an alternative stopping
rule, but the submitted candidate does not improve over the incumbent and is
therefore rejected. A later branch introduces a metric placement skill and
reaches 100\% formal success, after which the same frozen harness is
re-evaluated and again obtains 100\%. This trajectory illustrates that
individual development attempts need not improve monotonically: VISTA can
retain the best-so-far harness while continuing to explore alternative
hypotheses and reuse evidence from unsuccessful branches.

\textbf{Acquired Harness Mechanisms.}
We examine the artifacts produced during harness development to understand what is accumulated through optimization.
The revisions span perception interfaces, state verification, action composition, and the allocation of control between executable skills and the frozen VLA.
These categories are not mutually exclusive: a single revision may change both how an object is localized and how the outcome of the subsequent manipulation is verified.
For example, across the developed harnesses, the optimizer introduces observation-dependent manipulation targets, explicit grasp and execution-state verification, bounded recovery procedures, and programmatic action sequences that complement VLA control.

\begin{figure*}[t]
\centering
\includegraphics[width=\linewidth]{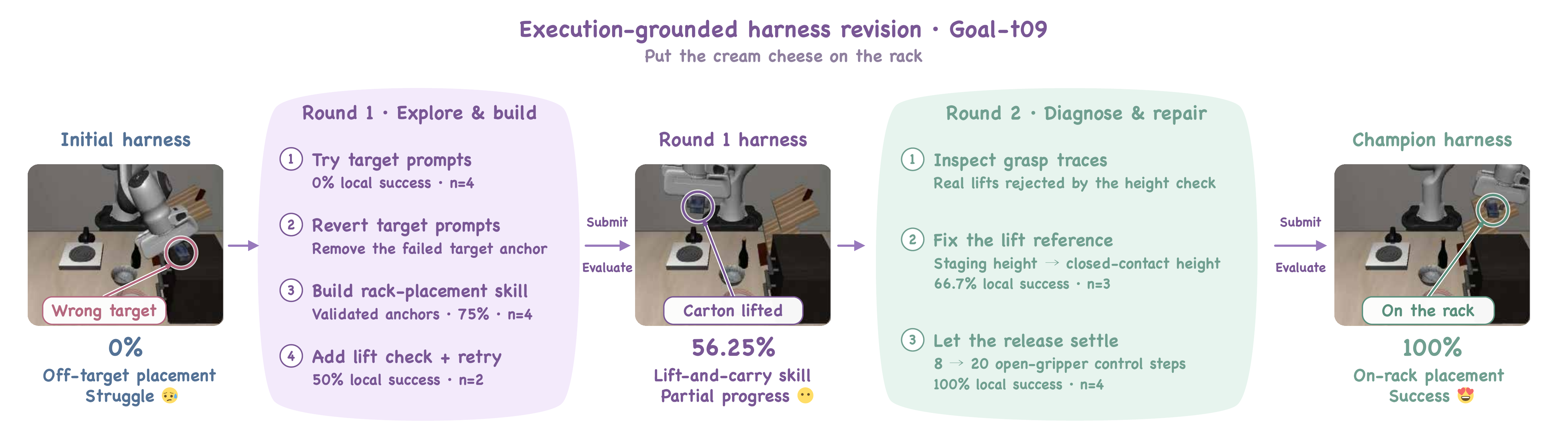}

\caption{
\textbf{A representative harness development process on Goal-t09.}
The initial harness achieves \rev{0.0}\rev{\%} \rev{success}.
The first development round adds lift verification, a bounded re-localization and grasp retry, and an explicit transport-and-release procedure, improving performance to \rev{56.3}\rev{\%}.
Analysis of the remaining failures then leads to a revision of the lift-state interpretation and release behavior, after which the final harness achieves \rev{100.0}\rev{\%} \rev{in autonomous evaluation}.
}
\label{fig:development-pro}
\vspace{-0.5cm}
\end{figure*}

Figure~\ref{fig:development-pro} shows a representative example from Goal-t09.
The initial harness succeeds in \rev{0.0}\rev{\%} \rev{of} episodes.
In the first development round, the meta-agent analyzes the failed trajectories and makes several concrete changes: it requires a verified lift before transport, re-segments the carton before a bounded grasp retry, and introduces an explicit transport-and-release procedure for placing the carton onto the rack.
The resulting harness reaches \rev{56.3}\rev{\%}.
The remaining failures expose a mismatch in the lift verification logic: lift was measured relative to an approach pose rather than the closed-contact state from which the carton was actually lifted.
The second round therefore revises the detector to measure displacement over a continuous closed-contact segment and resets the accumulated evidence when contact is lost, together with further adjustments to the release behavior.
After these revisions, the harness reaches \rev{100.0}\rev{\%} \rev{in autonomous evaluation}.
The accepted changes from the first round remain part of the harness in the second round, so later development builds directly on mechanisms acquired earlier rather than restarting from the initial system.

\section{Conclusion}

We introduced VISTA, a framework for improving embodied agents by evolving their operational harness while keeping the underlying model \rev{parameters} fixed. Rather than treating completed rollouts as a static source of feedback, VISTA enables a meta-agent to actively acquire diagnostic evidence, revise executable harness components, and test intermediate solutions before submitting a frozen candidate for autonomous evaluation. Across LIBERO-Pro\rev{, RoboCasa,} and \rev{RoboTwin}, this process progressively improves task success over successive development rounds. These results demonstrate that active, execution-grounded harness development provides a practical mechanism for converting embodied interaction into persistent improvements in agent behavior.

\textbf{Limitations and Future Works.} While VISTA demonstrates substantial gains across simulated benchmarks, our current evaluations are conducted strictly within high-fidelity simulation environments. \rev{Physical} real-robot validation has not yet been performed. Real-world deployment inevitably introduces sensor noise, unpredictable actuation latency, and unmodeled contact dynamics.
In future work, we plan to validate VISTA on physical robotic platforms (e.g., Franka Emika arms and mobile manipulators), \rev{and evaluate sensing, control constraints,} and \rev{supervision needed for safe physical} exploration.

% 文献以 原内容稿 iclr2027_conference.bib 为准；模板自带 main.bib/paper.bib 不混用。
\bibliographystyle{plain}
\bibliography{iclr2027_conference}

\clearpage
\appendix
\section*{Appendix}

\setcounter{figure}{0}
\renewcommand{\thefigure}{A\arabic{figure}}
\renewcommand{\theHfigure}{A\arabic{figure}}

\setcounter{table}{0}
\renewcommand{\thetable}{A\arabic{table}}
\renewcommand{\theHtable}{A\arabic{table}}

\section{Additional Experimental Details}
\label{sec:appendix:details}

\subsection{Infrastructure}
\label{sec:appendix:infrastructure}

\begin{figure}[t]
    \centering
    \includegraphics[width=\linewidth]{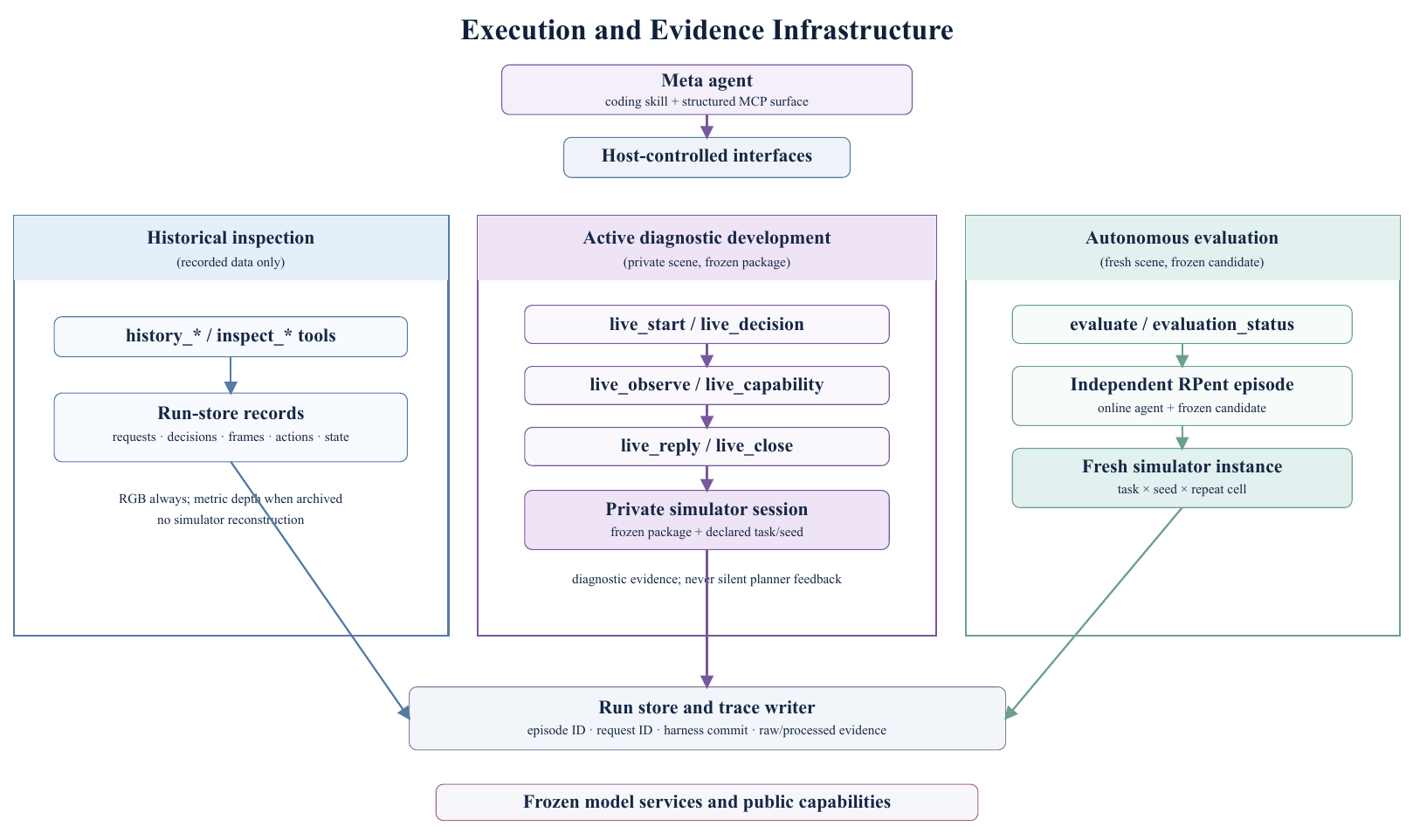}
    \caption{\textbf{Execution and evidence infrastructure.} The meta-agent communicates through a structured MCP interface. Historical inspection reads completed records; live development interacts with a private scene created from a frozen harness snapshot; and autonomous evaluation runs fresh episodes with no meta-agent intervention. All lanes use request, episode, and harness identities. Evidence is written to the run store, while package freezing and formal selection remain host-controlled. Historical depth is available only for trace modes that archive metric depth and calibration.}
    \label{fig:execution-infrastructure}
\end{figure}

VISTA separates harness development from autonomous evaluation through three interfaces. The meta-agent can read previous execution records, edit a candidate package, run live diagnostics, and request autonomous tests. The host freezes package versions, starts simulator processes, records traces, and applies the final selection rule. Figure~\ref{fig:execution-infrastructure} shows the three execution paths. Historical inspection reads completed records and does not recreate a previous simulator state. Live development starts a private scene from a frozen harness snapshot. Autonomous evaluation starts fresh episodes with a frozen candidate and no meta-agent intervention. All paths write records with explicit episode, request, and harness identities.

A live scene is not hot-swapped when the worktree changes. To test a later revision, VISTA starts a new session with the new frozen package. Live observations are stored as diagnostic evidence and are not silently added to the planner's context. Historical depth is available only for trace modes that archive metric depth and calibration; \texttt{full\_rgb} records contain RGB but no historical depth.

\subsection{Meta-Agent Skill and Tool Surface}
\label{sec:appendix:meta-skill}

The meta-agent follows the \texttt{evolve-harness} skill. This skill describes how to inspect history, formulate a concrete hypothesis, revise the harness, use live sessions for diagnosis, and validate the resulting change. It is a development protocol; it does not change the planner or VLA weights. Table~\ref{tab:meta-tool-surface} summarizes the main interfaces exposed to the meta-agent.

\begin{table*}[t]
    \centering
    \small
    \setlength{\tabcolsep}{5pt}
    \caption{\textbf{Meta-agent interfaces in the VISTA implementation.} Tools are grouped by their role in harness development.}
    \label{tab:meta-tool-surface}

    \begin{tabular}{
        >{\RaggedRight\arraybackslash}p{0.20\linewidth}
        >{\RaggedRight\arraybackslash}p{0.33\linewidth}
        >{\RaggedRight\arraybackslash}p{0.37\linewidth}
    }
        \toprule
        Interface & Representative tools & Returned evidence or effect \\
        \midrule

        History inspection
        &
        \texttt{history\_summary},\newline
        \texttt{history\_episodes},\newline
        \texttt{history\_episode},\newline
        \texttt{history\_decisions},\newline
        \texttt{history\_decision},\newline
        \texttt{history\_timeline},\newline
        \texttt{history\_frame},\newline
        \texttt{inspect\_decision},\newline
        \texttt{inspect\_harness}
        &
        Versioned harness metadata, task/seed identities, exact planner
        decisions, tool arguments, raw and processed feedback, physical
        step ranges, images, and package contents.
        \\

        \addlinespace

        Package development
        &
        Coding-agent file tools,\newline
        \texttt{select\_parent},\newline
        \texttt{budget},\newline
        \texttt{submit}
        &
        Candidate edits, saved drafts, selected Git parents, budget state,
        package validation, commit identity, and submission notes.
        \\

        \addlinespace

        Active diagnostics
        &
        \texttt{live\_start},\newline
        \texttt{live\_decision},\newline
        \texttt{live\_observe},\newline
        \texttt{live\_capability},\newline
        \texttt{live\_reply},\newline
        \texttt{live\_close}
        &
        A private frozen-package session, decision-boundary observations,
        public capability probes, optional decision replacement, and
        diagnostic episode records.
        \\

        \addlinespace

        Autonomous testing
        &
        \texttt{evaluate},\newline
        \texttt{evaluation\_status}
        &
        A frozen candidate and fresh autonomous episodes for specified
        task--seed--repeat cells. Autonomous tests requested during development are diagnostic evidence only. Promotion is based on a separate host-initiated evaluation after submission.
        \\

        \bottomrule
    \end{tabular}
\end{table*}
\FloatBarrier

The harness package itself exposes a separate runtime surface to the online
agent. In the LIBERO reference package, this surface includes native file and
termination tools, image reading, an observation skill, and a VLA-step skill.
A package may additionally define a \texttt{Policy} with
\texttt{prepare\_request}, \texttt{process\_tool\_result}, and optional
\texttt{decide} hooks. These hooks operate inside the ordinary host planner
loop: they can transform the visible request or choose among bound tools, but
they do not replace the host loop or change the model weights and budgets.

The distinction between the two surfaces is deliberate. Meta tools control
development and inspection; harness tools determine what the online agent can
use during an autonomous episode. A live call made by the meta-agent therefore
does not imply that the corresponding operation was available to, or selected
by, the online agent in the formal evaluation.

\subsection{Git-Backed Harness History}
\label{sec:appendix:git-history}

Each frozen harness is stored as a Git commit. Ordinary revisions create
parent edges, and an explicit merge creates a commit with multiple parents.
The run store attaches evaluation requests, episode records, live diagnostics,
and selection decisions to the corresponding commit.

Rejected candidates remain reachable through retained references and may be
selected as later development bases. The incumbent used for formal comparison
is recorded separately from the parent selected for revision. Thus, Git
records code lineage, while the run store records the evidence and decisions
associated with each version. Reusing a file or an idea does not create a
Git ancestry edge.

\section{Additional Quantitative Results}

\subsection{Complete Per-Task Results}

Tables~\ref{tab:libero_per_task}\rev{,~\ref{tab:libero_s_per_task},} and~\ref{tab:robocasa_per_task}
report per-task \rev{success rates} on
the \rev{LIBERO-T suites, LIBERO-S suites,} and RoboCasa, respectively.
H0 denotes the unmodified harness, while Evolved denotes the retained champion.
\rev{Each task is} evaluated \rev{using eight scene seeds with two repetitions per seed}.

\paragraph{LIBERO-Pro.}
Table~\ref{tab:libero_per_task} shows improvements over H0 on all 40 tasks
across the four \rev{LIBERO-T} suites. The suite-average success rate increases from 19.0\%
to 98.0\% on Spatial, from 49.0\% to 100.0\% on Object, from 16.0\% to
99.0\% on Goal, and from 7.0\% to 89.0\% on LIBERO-10. The retained champions
achieve 100.0\% success in \rev{autonomous} evaluation on 34 of the
40 tasks, including every task in Object. The task-level breakdown also
reveals differences that are less apparent from the suite averages. Six tasks
remain below 100\% success: two in Spatial, one in Goal, and three in
LIBERO-10. In particular, LIBERO-10 T4 and T9 remain at 63.0\% and 44.0\%,
respectively, despite improving substantially over their zero-success initial
harnesses.

\begin{table*}[t]
\centering
\caption{\textbf{Per-task success rate (\%) on \rev{the four LIBERO-T suites of} LIBERO-Pro.} H0 denotes the unmodified harness, and Evolved denotes the per-task champion after evolution. $\Delta$ is the gain over H0 in percentage points.}
\label{tab:libero_per_task}
\footnotesize
\setlength{\tabcolsep}{4.2pt}
\renewcommand{\arraystretch}{1.08}
\begin{adjustbox}{max width=\textwidth}
\begin{tabular}{llrrrrrrrrrrr}
\toprule
Suite & Method & T0 & T1 & T2 & T3 & T4 & T5 & T6 & T7 & T8 & T9 & Avg. \\
\midrule
\multirow{3}{*}{Spatial}
& H0         & 0.0 & 94.0  & 0.0 & 0.0 & 0.0 & 56.0 & 19.0  & 0.0  & 13.0 & 6.0   & 19.0 \\
& Evolved    & 100.0 & 100.0 & 100.0 & 100.0 & 94.0 & 100.0 & 100.0 & 88.0 & 100.0 & 100.0 & 98.0 \\
& $\Delta$   & +100.0 & +6.0 & +100.0 & +100.0 & +94.0 & +44.0 & +81.0 & +88.0 & +88.0 & +94.0 & +79.0 \\
\midrule
\multirow{3}{*}{Object}
& H0         & 94.0  & 81.0  & 69.0 & 0.0 & 56.0 & 13.0 & 6.0  & 19.0 & 75.0  & 81.0  & 49.0 \\
& Evolved    & 100.0 & 100.0 & 100.0 & 100.0 & 100.0 & 100.0 & 100.0 & 100.0 & 100.0 & 100.0 & 100.0 \\
& $\Delta$   & +6.0  & +19.0 & +31.0 & +100.0 & +44.0 & +88.0 & +94.0 & +81.0 & +25.0 & +19.0 & +51.0 \\
\midrule
\multirow{3}{*}{Goal}
& H0         & 0.0 & 6.0   & 31.0 & 0.0 & 31.0  & 0.0 & 25.0 & 19.0 & 50.0 & 0.0 & 16.0 \\
& Evolved    & 100.0 & 100.0 & 88.0 & 100.0 & 100.0 & 100.0 & 100.0 & 100.0 & 100.0 & 100.0 & 99.0 \\
& $\Delta$   & +100.0 & +94.0 & +56.0 & +100.0 & +69.0 & +100.0 & +75.0 & +81.0 & +50.0 & +100.0 & +83.0 \\
\midrule
\multirow{3}{*}{LIBERO-10}
& H0         & 0.0  & 25.0 & 38.0  & 0.0 & 0.0 & 0.0 & 6.0   & 0.0  & 0.0  & 0.0 & 7.0 \\
& Evolved    & 100.0 & 100.0 & 100.0 & 88.0 & 63.0 & 100.0 & 100.0 & 100.0 & 100.0 & 44.0 & 89.0 \\
& $\Delta$   & +100.0 & +75.0 & +63.0 & +88.0 & +63.0 & +100.0 & +94.0 & +100.0 & +100.0 & +44.0 & +83.0 \\
\bottomrule
\end{tabular}
\end{adjustbox}

\vspace{2pt}
\begin{minipage}{\textwidth}
\end{minipage}
\end{table*}
\FloatBarrier

\rev{Table~\ref{tab:libero_s_per_task} reports the corresponding per-task results for the four LIBERO-S suites.}

\begin{table*}[t]
\centering
\ifreviewversion\color{red}\fi
\caption{\textbf{\rev{Per-task success rate (\%) on the four LIBERO-S suites of LIBERO-Pro.} H0 denotes the unmodified harness, and Evolved denotes the per-task champion after evolution. $\Delta$ is the gain over H0 in percentage points.}}
\label{tab:libero_s_per_task}
\footnotesize
\setlength{\tabcolsep}{4.2pt}
\renewcommand{\arraystretch}{1.08}
\begin{adjustbox}{max width=\textwidth}
\begin{tabular}{llrrrrrrrrrrr}
\toprule
Suite & Method & S0 & S1 & S2 & S3 & S4 & S5 & S6 & S7 & S8 & S9 & Avg. \\
\midrule
\multirow{3}{*}{Spatial-S}
& H0 & 81.3 & 0.0 & 93.8 & 0.0 & 0.0 & 31.3 & 0.0 & 6.3 & 87.5 & 0.0 & 30.0 \\
& Evolved & 100.0 & 100.0 & 100.0 & 100.0 & 100.0 & 100.0 & 100.0 & 100.0 & 100.0 & 100.0 & 100.0 \\
& $\Delta$ & +18.8 & +100.0 & +6.3 & +100.0 & +100.0 & +68.8 & +100.0 & +93.8 & +12.5 & +100.0 & +70.0 \\
\midrule
\multirow{3}{*}{Object-S}
& H0 & 25.0 & 37.5 & 6.3 & 0.0 & 0.0 & 37.5 & 25.0 & 12.5 & 0.0 & 0.0 & 14.4 \\
& Evolved & 100.0 & 100.0 & 100.0 & 100.0 & 100.0 & 100.0 & 100.0 & 100.0 & 100.0 & 93.8 & 99.4 \\
& $\Delta$ & +75.0 & +62.5 & +93.8 & +100.0 & +100.0 & +62.5 & +75.0 & +87.5 & +100.0 & +93.8 & +85.0 \\
\midrule
\multirow{3}{*}{Goal-S}
& H0 & 0.0 & 6.3 & 0.0 & 6.3 & 0.0 & 0.0 & 0.0 & 6.3 & 12.5 & 12.5 & 4.4 \\
& Evolved & 100.0 & 100.0 & 100.0 & 93.8 & 100.0 & 100.0 & 100.0 & 100.0 & 81.3 & 100.0 & 97.5 \\
& $\Delta$ & +100.0 & +93.8 & +100.0 & +87.5 & +100.0 & +100.0 & +100.0 & +93.8 & +68.8 & +87.5 & +93.1 \\
\midrule
\multirow{3}{*}{LIBERO-10-S}
& H0 & 0.0 & 0.0 & 0.0 & 0.0 & 0.0 & 0.0 & 0.0 & 0.0 & 0.0 & 0.0 & 0.0 \\
& Evolved & 93.8 & 87.5 & 100.0 & 0.0 & 100.0 & 100.0 & 100.0 & 81.3 & 18.8 & 87.5 & 76.9 \\
& $\Delta$ & +93.8 & +87.5 & +100.0 & 0.0 & +100.0 & +100.0 & +100.0 & +81.3 & +18.8 & +87.5 & +76.9 \\
\bottomrule
\end{tabular}
\end{adjustbox}
\end{table*}
\FloatBarrier

\paragraph{RoboCasa.}
Table~\ref{tab:robocasa_per_task} reports per-task \rev{success rates} on \rev{Composite Seen} and \rev{Composite Unseen}. On \rev{Composite Seen}, the evolved harness \rev{improves} success from \rev{16.4}\% to \rev{24.2}\%, with \rev{11} of the 16 tasks \rev{improving} over \rev{$H_0$. The largest gains occur on PackIdenticalLunches, PreSoakPan, and WashLettuce. On Composite Unseen, success increases from 4.7}\rev{\%} \rev{to 13.3}\rev{\%}\rev{; 11 tasks improve}, with the largest \rev{gains} on \rev{CategorizeCondiments} and \rev{BreadSelection}.

\begin{table*}[t]
\centering
\caption{\textbf{\rev{Per-task success rates (}\rev{\%}\rev{)} on \rev{the} RoboCasa composite suites.} $\pi_{0.5}$ \rev{is} the \rev{reference policy}, \rev{$H_0$ is the initial harness,} and Evolved \rev{is} the \rev{selected harness}. \rev{$\Delta$} is \rev{the gain over $H_0$} in \rev{percentage points}.}
\label{tab:robocasa_per_task}
\scriptsize
\renewcommand{\arraystretch}{1.05}

\begin{minipage}[t]{0.49\textwidth}
\centering
\textbf{Composite-Seen}

\vspace{2pt}

\resizebox{\linewidth}{!}{
\begin{tabular}{lcccc}
\toprule
Task & $\pi_{0.5}$ (\%) & H0 & Evolved & $\Delta$ \\
\midrule
DeliverStraw & 0.0 & \rev{6.3} & \rev{12.5} & +\rev{6.3} \\
GetToastedBread & 0.0 & \rev{0.0} & \rev{0.0} & +\rev{0.0} \\
KettleBoiling & 2.0 & \rev{6.3} & \rev{18.8} & +\rev{12.5} \\
LoadDishwasher & 2.0 & \rev{37.5} & \rev{43.8} & +\rev{6.3} \\
PackIdenticalLunches & 4.0 & \rev{0.0} & \rev{18.8} & +\rev{18.8} \\
PreSoakPan & 26.0 & \rev{25.0} & \rev{43.8} & +\rev{18.8} \\
PrepareCoffee & 2.0 & \rev{0.0} & \rev{6.3} & +\rev{6.3} \\
RinseSinkBasin & 4.0 & \rev{37.5} & \rev{43.8} & +\rev{6.3} \\
ScrubCuttingBoard & 2.0 & \rev{37.5} & \rev{43.8} & +\rev{6.3} \\
SearingMeat & 0.0 & \rev{0.0} & \rev{0.0} & +0.0 \\
SetUpCuttingStation & 10.0 & \rev{12.5} & \rev{12.5} & +0.0 \\
StackBowlsCabinet & 26.0 & \rev{37.5} & \rev{50.0} & +\rev{12.5} \\
SteamInMicrowave & 2.0 & \rev{18.8} & \rev{31.3} & +\rev{12.5} \\
StirVegetables & 6.0 & \rev{12.5} & \rev{12.5} & +0.0 \\
StoreLeftoversInBowl & 0.0 & \rev{6.3} & \rev{6.3} & +0.0 \\
WashLettuce & 22.0 & \rev{25.0} & \rev{43.8} & +\rev{18.8} \\
\midrule
\textbf{\rev{Average}} & \textbf{6.8} & \textbf{\rev{16.4}} & \textbf{\rev{24.2}} & \textbf{+\rev{7.8}} \\
\bottomrule
\end{tabular}
}
\end{minipage}
\hfill
\begin{minipage}[t]{0.49\textwidth}
\centering
\textbf{Composite-Unseen}

\vspace{2pt}

\resizebox{\linewidth}{!}{
\begin{tabular}{lcccc}
\toprule
Task & $\pi_{0.5}$ (\%) & H0 & Evolved & $\Delta$ \\
\midrule
ArrangeBreadBasket & 2.0 & \rev{12.5} & \rev{18.8} & +\rev{6.3} \\
ArrangeTea & 0.0 & \rev{0.0} & \rev{0.0} & +0.0 \\
BreadSelection & 2.0 & \rev{0.0} & \rev{18.8} & +\rev{18.8} \\
CategorizeCondiments & 0.0 & \rev{0.0} & \rev{37.5} & +\rev{37.5} \\
CuttingToolSelection & 8.0 & \rev{6.3} & \rev{12.5} & +\rev{6.3} \\
GarnishPancake & 0.0 & \rev{31.3} & \rev{31.3} & +0.0 \\
GatherTableware & 0.0 & \rev{0.0} & \rev{6.3} & +\rev{6.3} \\
HeatKebabSandwich & 0.0 & \rev{0.0} & \rev{0.0} & +0.0 \\
MakeIceLemonade & 0.0 & \rev{6.3} & \rev{6.3} & +0.0 \\
PanTransfer & 0.0 & \rev{0.0} & \rev{6.3} & +\rev{6.3} \\
PortionHotDogs & 0.0 & \rev{0.0} & \rev{6.3} & +\rev{6.3} \\
RecycleBottlesByType & 0.0 & \rev{0.0} & \rev{12.5} & +\rev{12.5} \\
SeparateFreezerRack & 0.0 & \rev{0.0} & \rev{0.0} & +0.0 \\
WaffleReheat & 0.0 & \rev{12.5} & \rev{25.0} & +\rev{12.5} \\
WashFruitColander & 2.0 & \rev{0.0} & \rev{12.5} & +\rev{12.5} \\
WeighIngredients & 0.0 & \rev{6.3} & \rev{18.8} & +\rev{12.5} \\
\midrule
\textbf{\rev{Average}} & \textbf{0.9} & \textbf{\rev{4.7}} & \textbf{\rev{13.3}} & \textbf{+\rev{8.6}} \\
\bottomrule
\end{tabular}
}
\end{minipage}

\end{table*}
\FloatBarrier

\subsection{Complete Evolution Details}

\begin{figure*}[t]
\centering
\begin{subfigure}[t]{0.49\textwidth}
  \includegraphics[width=\linewidth]{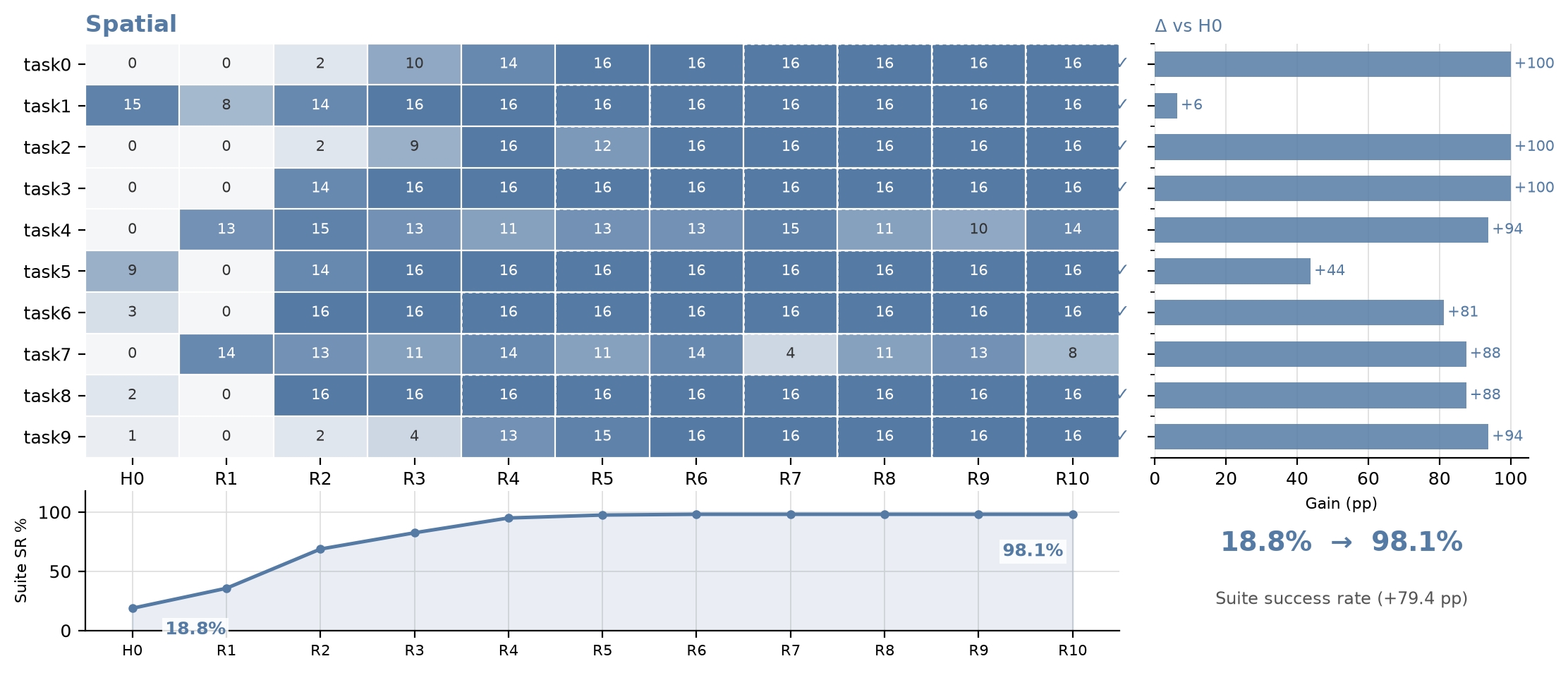}
  \caption{Spatial}
\end{subfigure}\hfill
\begin{subfigure}[t]{0.49\textwidth}
  \includegraphics[width=\linewidth]{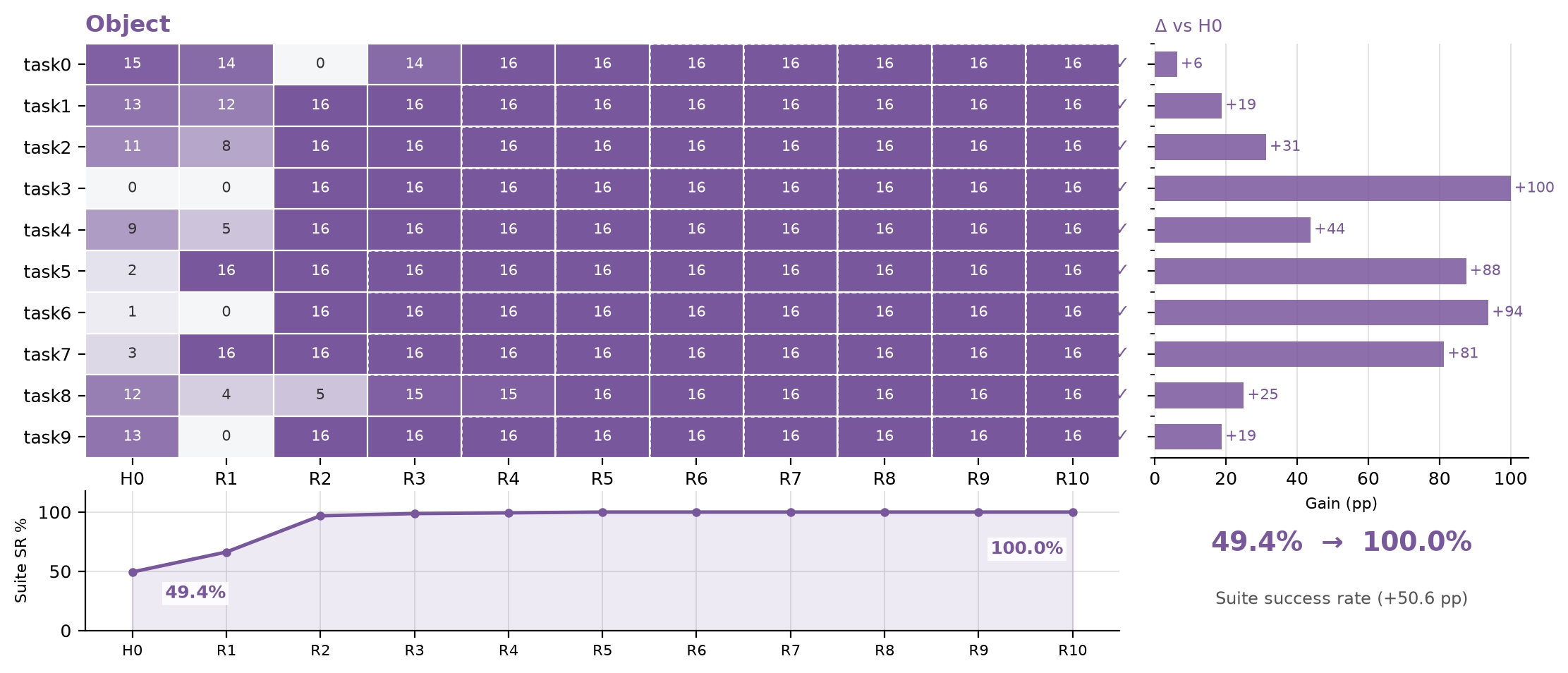}
  \caption{Object}
\end{subfigure}

\vspace{4pt}
\begin{subfigure}[t]{0.49\textwidth}
  \includegraphics[width=\linewidth]{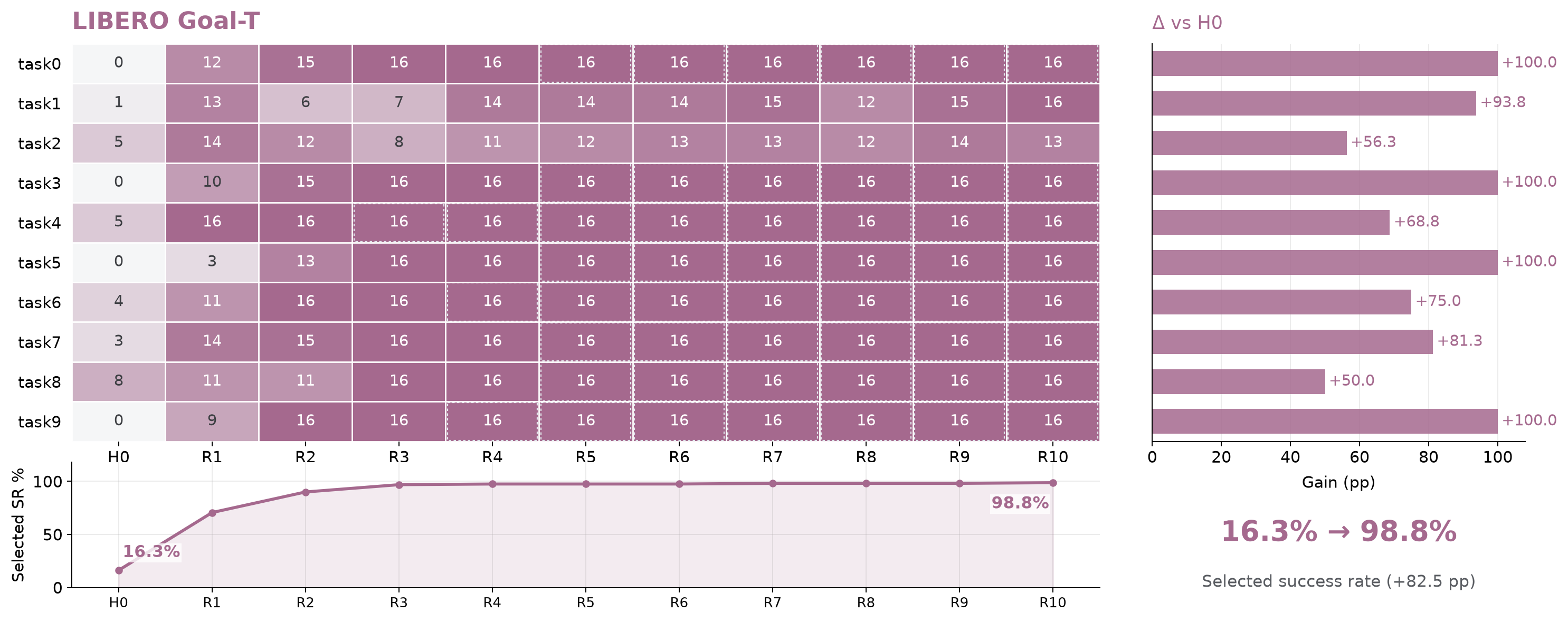}
  \caption{Goal}
\end{subfigure}\hfill
\begin{subfigure}[t]{0.49\textwidth}
  \includegraphics[width=\linewidth]{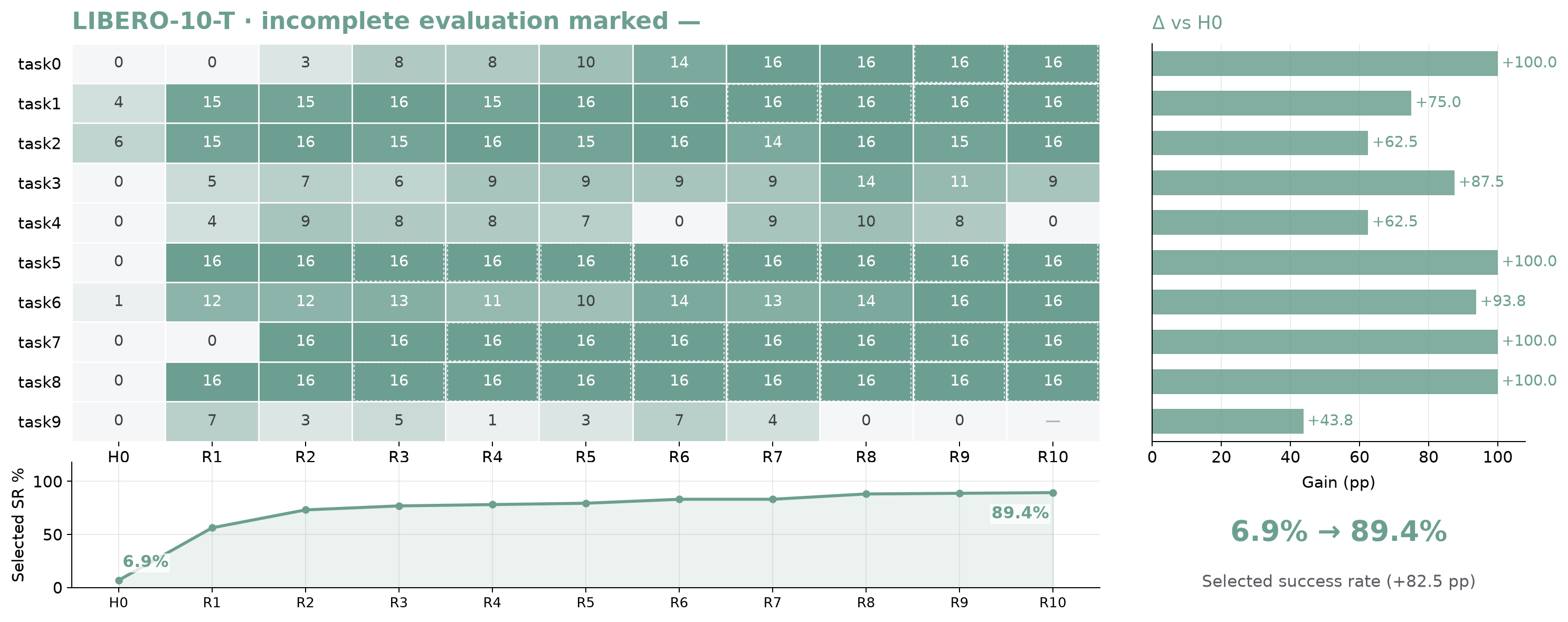}
  \caption{LIBERO-10}
\end{subfigure}
\caption{\textbf{Per-task evolution on \rev{the four LIBERO-T suites}.} Rows are tasks \rev{and} columns are \rev{development} rounds (H0\rev{: initial} harness). \rev{Solid cells report candidate} successes out of \rev{16 episodes. Dashed cells carry forward} the \rev{last retained result after development stops. An em dash marks an incomplete evaluation (LIBERO-10 task9, R10)}. Right: \rev{retained-harness} gain over H0. Bottom: \rev{aggregate retained scores}.}
\label{fig:evolution_heatmaps}
\end{figure*}

Figure~\ref{fig:evolution_heatmaps} presents the recorded development
trajectories for \rev{the} four \rev{LIBERO-T} suites.
The heatmaps show the performance of the candidate evaluated in each
round, whereas the curves below aggregate the best-so-far champion
scores across tasks.
The right-hand panels summarize the gain of each retained champion
over H0.
This distinction separates the quality of individual proposals from
the performance accumulated through candidate selection.

Across the four suites, the largest aggregate improvements occur
during the early rounds, followed by smaller, task-dependent gains.
Object reaches perfect development performance, while Spatial and
Goal approach saturation.
LIBERO-10 improves more gradually and retains a larger performance
gap at the reporting endpoint.
The per-task trajectories further show that progress is not synchronized
across tasks: some reach high success rates within the first few rounds,
whereas others continue to receive revisions throughout the displayed
development period.
For example, Goal T1 reaches $100.0\%$ only in R10.

Individual candidate scores are often non-monotonic, even when the
retained champion remains unchanged.
For example, Spatial T7 reaches $87.5\%$ in R1, but several subsequent
candidates obtain lower scores without replacing that champion.
A plateau in the aggregate curve therefore does not necessarily indicate
that development has stopped; it may instead reflect continued exploration
without a successful replacement.
Likewise, the non-decreasing champion curves follow from best-so-far
selection and do not imply that every proposed revision improves
performance.

\end{document}